\documentclass[a4paper,fleqn]{cas-dc}

\usepackage[numbers]{natbib}
\usepackage{amsmath}
\usepackage{mathtools}
\usepackage{array}
\usepackage{multirow}
\usepackage{multicol}
\usepackage[caption=false,font=normalsize,labelfont=sf,textfont=sf]{subfig}
\usepackage{url}
\usepackage{amsthm}

\usepackage[noabbrev]{cleveref}
\usepackage{amsfonts}

\usepackage{subfig}
\usepackage[autostyle]{csquotes}  
\usepackage{mwe}
\usepackage{physics}

\usepackage[linesnumbered,ruled,vlined]{algorithm2e}
\usepackage{algorithmic}

\usepackage{enumitem}
\setlist[enumerate]{leftmargin=*}
\setlist[itemize]{leftmargin=*}

\def\tsc#1{\csdef{#1}{\textsc{\lowercase{#1}}\xspace}}
\tsc{WGM}
\tsc{QE}
\tsc{EP}
\tsc{PMS}
\tsc{BEC}
\tsc{DE}

\begin{document}
\let\WriteBookmarks\relax
\def\floatpagepagefraction{1}
\def\textpagefraction{.001}

\title [mode = title]{Hybrid Variational Autoencoder for Time Series Forecasting}                      

\tnotetext[1]{Corresponding author}

%
\author[1]{Borui Cai}

\cormark[1]

\author[2]{Shuiqiao Yang}

\author[3]{Longxiang Gao}

\author[1]{Yong Xiang}

\affiliation[1]{organization={School of Information Technology, Deakin University},
    city={Burwood},
    postcode={3125}, 
    state={VIC},
    country={Australia}}

\affiliation[2]{organization={School of Computer Science and Engineering, University of New South Wales},
    city={Sydney},
    postcode={2032}, 
    state={NSW},
    country={Australia}}

\affiliation[3]{organization={Qilu University of Technology (Shandong Academy of Sciences)},
    city={Jinan},
    country={China}}

\nonumnote{\textit{Email addresses}: b.cai@deakin.edu.au (B.Cai); \newline
shuiqiao.yang@unsw.edu.au (S.Yang); 
gaolx@sdas.org (L.Gao); \newline
yong.xiang@deakin.edu.au (Y.Xiang)
  }

\begin{abstract}
Variational autoencoders (VAE) are powerful generative models that learn the latent representations of input data as random variables. Recent studies show that VAE can flexibly learn the complex temporal dynamics of time series and achieve more promising forecasting results than deterministic models. However, a major limitation of existing works is that they fail to jointly learn the local patterns (e.g., seasonality and trend) and temporal dynamics of time series for forecasting. Accordingly, we propose a novel hybrid variational autoencoder (HyVAE) to integrate the learning of local patterns and temporal dynamics by variational inference for time series forecasting. Experimental results on four real-world datasets show that the proposed HyVAE achieves better forecasting results than various counterpart methods, as well as two HyVAE variants that only learn the local patterns or temporal dynamics of time series, respectively.
\end{abstract}


\maketitle

\section{Introduction}
\label{sect:1}
Time series forecasting aims at learning the generation process of time series and uses previously observed samples to predict future values \cite{informer}. Accurate forecasting is essential and can help with the success of many applications/businesses. For example, an electricity company can design effective energy policies in advance by predicting the future energy consumption \cite{electricity}; a corporation can minimize its investment risk if the future stock prices are accurately predicted \cite{financial}.

Time series forecasting has been studied in the literature for decades, but to date, it remains a challenging and active research problem due to the complexity of time series. Classical time series forecasting methods, including autoregressive models (AR), moving average models (MA), and autoregressive integrated moving average models (ARIMA) \cite{arima}, predict future values by assuming they have linear relationships with observed values; however, this simplification normally leads to unsatisfactory results for complex real-world time series. With the booming of deep learning techniques, deep neural networks (DNN) are widely used to tackle time series forecasting problems. Unlike classical models, DNNs are flexible non-linear models that can capture the temporal information of time series for forecasting \cite{deepsurvey}. Convolutional neural networks (CNN) \cite{cnn} and recurrent neural networks (RNN) \cite{rnn} are two types of DNN widely adopted for time series forecasting. CNN captures salient local patterns of short time series subsequences/segments (e.g., seasonality \cite{season} and trend \cite{cnnlstm}), while RNN learns long-term or mid-term temporal dynamics/dependencies of the entire time series \cite{lstm}. In fact, many works capture both types of temporal information by proposing hybrid DNN models and obtaining more accurate forecasting results \cite{cnnlstm,tcnlstm}. For example, the researchers \cite{dna} adopts a hybrid neural network, which stacks CNN with RNN, for DNA sequence prediction. Specifically, CNN can capture short and recurring sequence motifs, which represent biological function units in a DNA sequence. RNN, i.e., long short-term memory (LSTM) \cite{lstm}, is stacked with the output of CNN to learn the spatial arrangement of these motifs. 

However, these DNN-based models cannot capture temporal information from time series with high accuracy since they are sensitive to small perturbations on time series \cite{overfitting}. Recent works refer to variational autoencoder (VAE) \cite{vae}, which is a type of deep generative model, to learn representations of time series as latent random variables and obtain improved results \cite{vrnn}. Compared with directly fitting the exact values of time series, the latent random variables learned by VAE represent the generation process of time series and thus can more accurately capture essential temporal information of time series \cite{deep}.
Based on this, existing methods learn either local seasonal-trend patterns \cite{last} or temporal dynamics \cite{vrnn}; but to date, there is no VAE model that can jointly capture both information for time series forecasting.

In this paper, we bridge this gap by proposing a novel hybrid variational autoencoder (HyVAE) method for time series forecasting. HyVAE follows the variational inference \cite{vae} to jointly learn local patterns and temporal dynamics of time series. To achieve this goal, HyVAE is designed based on two objectives: 1) capturing local patterns by encoding time series subsequences into latent representations; 2) learning temporal dynamics through the temporal dependencies among latent representations of different time series subsequences. HyVAE integrates the two objectives following the variational inference.
Extensive experiments conducted on four real-world time series datasets show that HyVAE can improve the time series forecasting accuracy over strong counterpart methods. 
The contributions of this paper are summarized as follows:
\begin{itemize}
\item[--] We propose a novel hybrid variational autoencoder (HyVAE) for time series forecasting. HyVAE derives an objective following variational inference to integrate the learning of local patterns and temporal dynamics of time series, thereby improving the accuracy of forecasting.
\item[--] We conduct comprehensive experiments on four real-world datasets to demonstrate the effectiveness of the proposed HyVAE method, and the results show that HyVAE achieves better forecasting accuracy than strong counterpart methods.
\end{itemize}

The rest of this paper is organized as follows. The related works are reviewed in Section \ref{sect:2}. The preliminary knowledge is introduced in Section \ref{sect:3}. The proposed method is detailed in Section \ref{sect:4}, and is evaluated in Section \ref{sect:5}. The paper is summarized in Section \ref{sect:6}.

\section{Related Work}
\label{sect:2}
In this section, we briefly review time series forecasting methods and VAE-related forecasting approaches.

\subsection{Time series forecasting}
Classical auto-regressive model (AR) predicts by the linear aggregation of past time series values and a stochastic term (e.g., white noise). ARIMA extends AR to non-stationary time series by incorporating moving average (MA) and differencing. Other statistical models, such as linear regression \cite{hybrid} and support vector regression \cite{svr}, enhances the model capacity but still have limited expressiveness. DNNs are flexible non-linear models and are widely used for time series forecasting in recent years. Specifically, RNNs memorize historical information with feedback loops and can conveniently learn the temporal dynamics of time series. Long short-memory network (LSTM) \cite{lstm} is a typical RNN that alleviates gradient vanishing with forget gates, and that enables the learning of long-term temporal dynamics for time series. Other types of RNN, e.g., GRU \cite{gru}, and Informer \cite{informer}, which uses the attention mechanism \cite{attention1}, are also used to improve the effectiveness of different forecasting scenarios. In addition, CNNs \cite{tcn} are further adopted to capture local patterns of time series (such as seasonality \cite{season} and trends \cite{cnnlstm}). Many works stack CNN and RNN to learn both the local patterns and the temporal dynamics for challenging forecasting problems; for example, combining multi-layer one-dimensional CNNs with bi-directional LSTM for air quality forecasting \cite{cnnlstm} and DNA sequence forecasting \cite{dna}; integrating a Savitzky–Golay filter (to avoid noise) and a stacked TCN-LSTM for traffic forecasting \cite{tcnlstm}.

\subsection{Variational autoencoder-based forecasting}
Variational autoencoder (VAE) \cite{vae} is a powerful deep generative model that encodes the input data as latent random variables, instead of deterministic values. To enhance the flexibility of VAE (learns independent latent random variables), follow-up methods introduce extra dependencies among the latent random variables. For example, ladder variational autoencoder \cite{lvae} specifies a top-down hierarchical dependency among the latent random variables, fully-connected variational autoencoder \cite{fcvae} includes all possible dependencies among variables, and graph variational autoencoder \cite{gvae} automatically learns an acyclic dependency graph. Due to the high flexibility, it is introduced to time series forecasting \cite{predictionvae1}. To improve the performance of the vanilla VAE, VRNN \cite{vrnn} introduces an RNN as the backbone to capture the long-term temporal dynamics of time series. LaST \cite{last} develops disentangled VAE to learn dissociated seasonality and trend patterns of time series for forecasting.
The proposed HyVAE is different from existing methods as it integrates the learning of both local patterns and the temporal dynamic for time series forecasting.

\section{Preliminaries}
\label{sect:3}
In this section, we first define the problem and then introduce the preliminary knowledge of VAE.

\subsection{Notation and problem statement}
A scalar is denoted as a lowercase character, a vector is denoted as a bold lowercase character, and a matrix is denoted as an uppercase character. A time series is denoted as $\boldsymbol{s}=\{s_{1},s_{2},...,s_{m},s_{m+1},...,s_{m+n}\}$, the time series forecasting problem is defined as determining $\{s_{m+1},...,s_{m+n}\}$ with known $\{s_{1},s_{2},...,s_{m}\}$, where $n$ is the step of forecasting. For the convenience, we denote $\boldsymbol{y}=\{s_{m+1},...,s_{m+n}\}$, and the forecasting problem can be formulated as $\hat{\boldsymbol{y}}=f(s_{1},s_{2},...,s_{m})$, where $\hat{\boldsymbol{y}}$ is the predicted values for $\boldsymbol{y}$. The error of forecasting is measured as follows:
\begin{equation}
Err(\boldsymbol{y},\hat{\boldsymbol{y}})=\frac{1}{n}\sum_{i=1}^{n}(y_{i}-\hat{y}_{i})^{2},
\label{eq:err}
\end{equation}

Time series subsequence is denoted as $\boldsymbol{x}_{t}=\{s_{t},..,s_{t+l-1}\}$, where $l$ is the length. Time series subsequence contains contextual information that expresses local patterns \cite{cnnlstm}, and thus we use subsequences in the forecasting task. Following \cite{ticc}, we obtain a series of $l$ length subsequences from time series, using a sliding window. Time series represented by subsequences is denoted as $\{\boldsymbol{x}^{1},...,\boldsymbol{x}^{T}\}$, where $T=m-l+1$ is the number of its subsequences. Thus, the forecasting problem becomes $\hat{\boldsymbol{y}}=f(\boldsymbol{x}^{\leq T})=f(\boldsymbol{x}^{1},...,\boldsymbol{x}^{T})$.

\begin{table}[!t]
\renewcommand{\arraystretch}{1.3}
\caption{Summary of notations.}
\label{tab:note}
\centering
\begin{tabular}{cl}
\hline
Notation & Description \\
\hline
$\boldsymbol{s}$ & time series\\
$\boldsymbol{y}$ & ground truth future values, $\{s_{m+1},...,s_{m+n}\}$\\
$\boldsymbol{\hat{y}}$ & predicted future values\\
$\boldsymbol{x}$ & time series subsequence, $\{s_{t},..,s_{t+l-1}\}$\\
$\boldsymbol{z}$ & latent representations learnt with VAE\\
$\boldsymbol{h}$ & hidden states learned with RNN (i.e., GRU)\\
$\boldsymbol{L}$ & $ladder\ size$ for subsequence encoding\\
$\mathcal{N}(\mu,\sigma)$ & Gaussian distribution\\
$KL(q(x)||p(x))$ & KL divergence from $q(x)$ to $p(x)$\\
\hline
\end{tabular}
\end{table}

\subsection{Variational autoencoder}
\begin{figure}[htbp]
\centering
\subfloat{\includegraphics[width=3.3in]{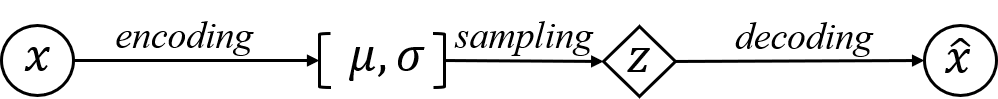}} \\
\caption{The framework of VAE. VAE encodes the input ($\boldsymbol{x}$) into the latent random variables (as Gaussian distributions). Then, $z$ is sampled from the distribution of latent random variables to reconstruct the input ($\hat{\boldsymbol{x}}$).}
\label{fig:vae}
\end{figure}

Variational autoencoder (VAE) \cite{vae} is an unsupervised generative learning model that learns the latent representation of the input data as random variables. Similar to the conventional autoencoder \cite{autoencoder}, VAE has an encoding process that encodes the input into latent representations, and a decoding process that reconstructs the original input with the learned representations. We show the process of VAE in Figure \ref{fig:vae}. 

VAE learns the generative model as $p(\boldsymbol{x},\boldsymbol{z}) = p(\boldsymbol{x}|\boldsymbol{z})p(\boldsymbol{z})$, where $\boldsymbol{x}$ is the input data and $\boldsymbol{z}$ is its latent representations. The prior of $\boldsymbol{z}$, $p(\boldsymbol{z})$, is normally defined as a multivariate Gaussian distribution, i.e., $\boldsymbol{z}\sim \mathcal{N}(\boldsymbol{0},I)$; we denote that as $p(\boldsymbol{z})= \mathcal{N}(\boldsymbol{z}|\boldsymbol{0},I)$ for convenience. The posterior $p(\boldsymbol{z}|\boldsymbol{x})$ normally can be an arbitrary non-linear non-Gaussian distribution and thus is intractable. To resolve that, VAE approximates the posterior with $q(\boldsymbol{z}|\boldsymbol{x})=\mathcal{N}(\boldsymbol{z}|\boldsymbol{\mu}(\boldsymbol{x}),\boldsymbol{\sigma}(\boldsymbol{x}))$, where mean and variance are determined by $\boldsymbol{x}$. Then, VAE defines the learning problem as the maximum likelihood estimation of $\log{p(\boldsymbol{x})}$, which can be formulated as:
\begin{equation}
\log{p(\boldsymbol{x})}=KL\Big(q(\boldsymbol{z}|\boldsymbol{x})||p(\boldsymbol{z}|\boldsymbol{x})\Big)+\ell,
\label{eq:px}
\end{equation}
where the first term is the KL divergence between the approximated posterior and the true posterior. Specifically, the KL divergence of two distributions $q(\boldsymbol{x})$ and $p(\boldsymbol{x})$ measures their similarity and is defined as:
\begin{equation}
KL\Big(q(\boldsymbol{x})||p(\boldsymbol{x})\Big)=\sum_{\boldsymbol{x}}q(\boldsymbol{x})\frac{q(\boldsymbol{x})}{p(\boldsymbol{x})}=\mathbb{E}_{q(\boldsymbol{x})}\frac{q(\boldsymbol{x})}{p(\boldsymbol{x})}.
\label{eq:kl}
\end{equation}
In Eq. (\ref{eq:px}), since $p(\boldsymbol{z}|\boldsymbol{x})$ is intractable and KL divergence is non-negative, maximizing $\log{p(\boldsymbol{x})}$ is achieved by maximizing $\ell$, which is the evidence lower bound (ELBO) of $\log{p(\boldsymbol{x})}$ defined as follows:
\begin{equation}
\ell=E_{q(\boldsymbol{z}|\boldsymbol{x})}\log{p(\boldsymbol{x}|\boldsymbol{z})}-KL\Big(q(\boldsymbol{z}|\boldsymbol{x})||p(\boldsymbol{z})\Big),
\label{eq:elbovae}
\end{equation}
The first term in $\ell$ maximizes the conditional probability of $\boldsymbol{x}$ given the latent representation $\boldsymbol{z}$ and can be seen as the reconstruction loss, while the second term minimizes the difference between the prior and the approximated posterior.

\section{The Proposed Method}
\label{sect:4}
In this section, we first provide an overview of the proposed hybrid variational autoencoder (HyVAE) method and then elaborate on its details.   

\subsection{Overview of HyVAE}
Inspired by existing deterministic deep neural models, we propose a novel generative hybrid variational autoencoder (HyVAE) model for time series forecasting. HyVAE jointly learns the local patterns from time series subsequences and the temporal dynamics among time series subsequences. To achieve that, HyVAE is derived based on variational inference to integrate two processes: 1) the encoding of time series subsequences, which captures local patterns; and 2) the encoding of entire time series, which learns temporal dynamics among time series subsequences. In the following content, we separately detail the encoding of time series subsequences and the encoding of the entire time series, respectively, and then explain the integration of these two processes for time series forecasting.

\subsection{Encoding of time series subsequence}
\label{section:sub}
As discussed in Section \ref{sect:1}, many existing models have shown that learning the local patterns can effectively improve time series forecasting \cite{cnnlstm}. To capture the flexible local patterns, we encode time series subsequences as latent random variables, rather than deterministic values.

The conventional VAE encodes a subsequence ($\boldsymbol{x}^{t}$) into latent representations as independent random variables ($\boldsymbol{z}^{t}$), which follow a multivariate Gaussian distribution, i.e., $p(\boldsymbol{z}^{t})=\mathcal{N}(\boldsymbol{z}^{t}|\boldsymbol{\mu}^{t},\boldsymbol{\sigma}^{t})$ and $\boldsymbol{\sigma}^{t}$ is the diagonal of a diagonal covariance matrix. However, the values of a time series subsequence are normally not independent and have a causal relationship (e.g., autoregressive); that is, the independent latent random variables cannot properly preserve the meaningful causal information \cite{tcnlstm} within a subsequence. Inspired by ladder variational autoencoder (LVAE) \cite{ladder1}, we enforce a hierarchical dependency among the latent random variables to capture the causal information in a subsequence. 

\begin{figure}[htbp]
\centering
\subfloat{\includegraphics[width=3.2in]{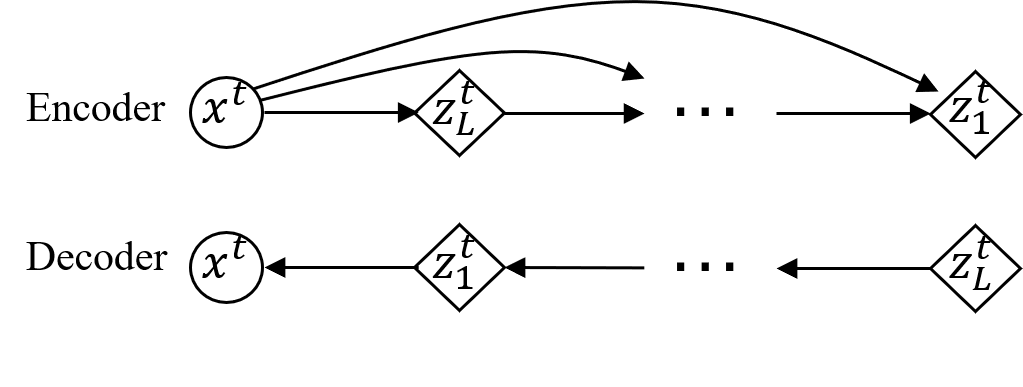}} \\
\caption{The encoder (inference process) and the decoder (generative process) of subsequence encoding.}
\label{fig:lvae}
\end{figure}

We separate the latent random variables $\boldsymbol{z}^{t}$ into $L$ groups/ladders ($L$ is the $ladder\ size$), i.e, $\{\boldsymbol{z}^{t}_{1},...,\boldsymbol{z}^{t}_{L}\}$, and the groups have a top-down hierarchical dependency (from $1$ to $L$). We illustrate the encoding and decoding process of subsequence encoding in Fig. \ref{fig:lvae}, in which the top row is the encoding process and the bottom row is the decoding process. For the convenience of implementation, we adopt the same top-down dependency among the latent random variables ($\boldsymbol{z}^{t}_{1}\xrightarrow[]{}...\xrightarrow[]{}\boldsymbol{z}^{t}_{L}$) in the encoding and decoding processes. Based on this, the prior distribution of $\boldsymbol{z}^{t}$ can be factorized as:
\begin{equation}
\begin{split}
p(\boldsymbol{z}^{t})&=p(\boldsymbol{z}^{t}_{L})\prod_{i=1}^{L-1}p(\boldsymbol{z}^{t}_{i}|\boldsymbol{z}^{t}_{i+1}),
\\
p(\boldsymbol{z}^{t}_{i}|\boldsymbol{z}^{t}_{i+1})&=\mathcal{N}(\boldsymbol{z}^{t}_{i}|\boldsymbol{\mu_{i}}^{t}(\boldsymbol{z}_{i+1}^{t}),\boldsymbol{\sigma}_{i}^{t}(\boldsymbol{z}_{i+1}^{t})),
\end{split}
\label{eq:laveprior}
\end{equation}
where $\{\boldsymbol{\mu}(\star),\boldsymbol{\sigma}(\star)\}=\varphi(\star)$ and we implement $\varphi(\star)$ as a multilayer perceptron (MLP). 
By changing the size of dependency ($L$, the $ladder\ size$), we can regulate how well causal information is preserved, and no causal information when $L=1$ (i.e., all latent random variables are independent).
Based on this, the generative model of subsequence encoding can further be factorized as follows:
\begin{equation}
\begin{split}
p(\boldsymbol{x}^{t},\boldsymbol{z}^{t})=p(\boldsymbol{x}^{t}|\boldsymbol{z}_{1}^{t})p(\boldsymbol{z}_{L}^{t})\prod_{i=1}^{L-1}p(\boldsymbol{z}_{i}^{t}|\boldsymbol{z}_{i+1}^{t}),
\end{split}
\label{eq:lavegenetion}
\end{equation}
where $p(\boldsymbol{x}^{t}|\boldsymbol{z}^{t}_{1})=\mathcal{N}(\boldsymbol{x}^{t}|\boldsymbol{\mu}_{i}^{t}(\boldsymbol{z}^{t}_{1}),\boldsymbol{\sigma}_{i}^{t}(\boldsymbol{z}^{t}_{1}))$.
This hierarchical dependency ensures the latent random variables have sufficient flexibility to model the complex local patterns of subsequences. Since the posterior $p(\boldsymbol{z}^{t}|\boldsymbol{x}^{t})$ is intractable, $q(\boldsymbol{z}^{t}|\boldsymbol{x}^{t})$ is used as an approximation. Meanwhile, to avoid $\{z_{L}^{t},...,z_{1}^{t}\}$ converging to arbitrary variables, they all depend on $x^{t}$ in the inference model similar to \cite{lvae} as follows:
\begin{equation}
\begin{split}
q(\boldsymbol{z}^{t}|\boldsymbol{x}^{t})&=q(\boldsymbol{z}_{L}^{t}|\boldsymbol{x}^{t})\prod_{i=1}^{L-1}q(\boldsymbol{z}_{i}^{t}|\boldsymbol{z}_{i+1}^{t},\boldsymbol{x}^{t}),
\\
q(\boldsymbol{z}_{L}^{t}|\boldsymbol{x}^{t})&=\mathcal{N}(\boldsymbol{z}_{L}^{t}|\boldsymbol{\mu}_{i}^{t}(\boldsymbol{x}^{t}),\boldsymbol{\sigma}_{i}^{t}(\boldsymbol{x}^{t})),
\\
q(\boldsymbol{z}_{i}^{t}|\boldsymbol{z}_{i+1}^{t},\boldsymbol{x}^{t})&=\mathcal{N}(\boldsymbol{z}_{i}^{t}|\boldsymbol{\mu}_{i}^{t}(\boldsymbol{z}_{i+1}^{t},\boldsymbol{x}^{t}),\boldsymbol{\sigma}_{i}^{t}(\boldsymbol{z}_{i+1}^{t},\boldsymbol{x}^{t})),
\end{split}
\label{eq:laveinference}
\end{equation}
where $\{\boldsymbol{\mu}(\star,\star),\boldsymbol{\sigma}(\star,\star)\}=\varphi([\star;\star])$ and $[;]$ is the concatenation operation.

\begin{figure*}[htbp]
\centering
\subfloat[Prior]{\includegraphics[width=1.7in]{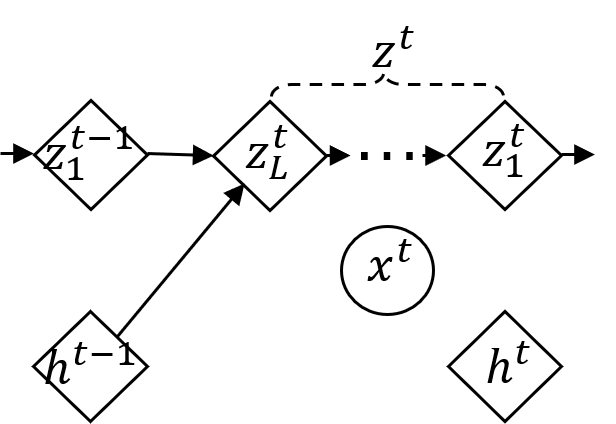}}\hspace{0.05in}
\subfloat[Recurrence]{\includegraphics[width=1.55in]{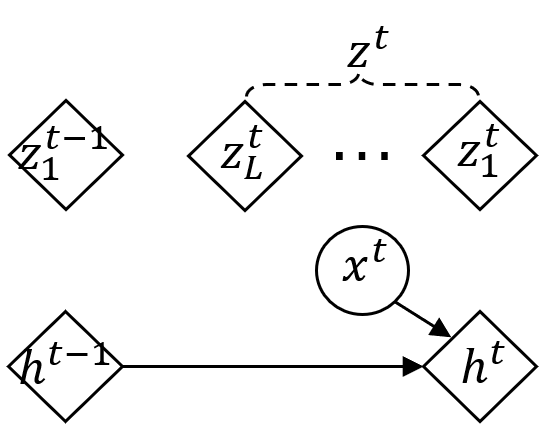}} \hspace{0.1in}
\subfloat[Inference]{\includegraphics[width=1.7in]{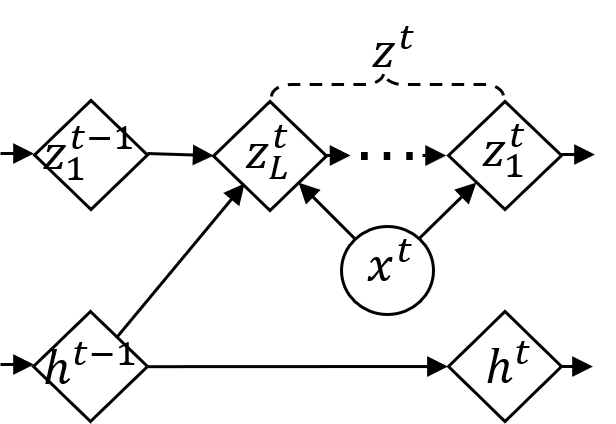}} \hspace{0.05in}
\subfloat[Generation]{\includegraphics[width=1.55in]{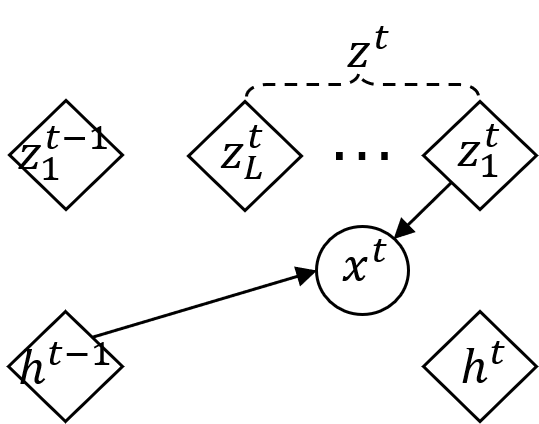}} \\
\caption{The illustration of HyVAE; (a) shows the prior defined by Eq. (\ref{eq:newprior}); (b) is the recurrent updating of GRU hidden states in Eq. (\ref{eq:tvaerecurrent}); (c) shows the inference operation in Eq. (\ref{eq:newinference}); and (d) represents the generation operation in Eq. (\ref{eq:newgeneration}).}
\label{fig:rvae}
\end{figure*}

\subsection{Encoding of entire time series}
From the global perspective, we encode all time series subsequences $\{\boldsymbol{x}^{1},...,\boldsymbol{x}^{T}\}$ as $\{\boldsymbol{z}^{1},...,\boldsymbol{z}^{T}\}$ to learn the temporal dynamics of entire time series. Since time series subsequences are normally not independent across different time stamps, we first impose a temporal dependency for consecutive subsequences (e.g., $p(\boldsymbol{z}^{t},\boldsymbol{z}^{t-1})=p(\boldsymbol{z}^{t}|\boldsymbol{z}^{t-1})p(\boldsymbol{z}^{t-1})$). In addition, we capture long-term temporal dependency with other subsequences by hidden states of a recurrent neural network, i.e., gated recurrent unit (GRU) \cite{gru}. Therefore, we have the following derivation:
$p(\boldsymbol{z}^{t}|\boldsymbol{z}^{<t})$ can be derived as follows:
\begin{equation}
\begin{split}
p(\boldsymbol{z}^{t}|\boldsymbol{z}^{<t})&=p(\boldsymbol{z}^{t}|\boldsymbol{z}^{t-1},\boldsymbol{h}^{t-1}),
\\
p(\boldsymbol{z}^{t}|\boldsymbol{z}^{t-1},\boldsymbol{h}^{t-1})&=\mathcal{N}(\boldsymbol{z}^{t}|\boldsymbol{\mu}_{i}^{t}(\boldsymbol{z}^{t-1},\boldsymbol{h}^{t-1}),\boldsymbol{\sigma}_{i}^{t}(\boldsymbol{z}^{t-1},\boldsymbol{h}^{t-1})),
\end{split}
\label{eq:tvaepriortmp}
\end{equation}
where $\boldsymbol{h}$ is the hidden state and is obtained by:
\begin{equation}
\boldsymbol{h}^{t}=\text{GRU}(\boldsymbol{h}^{t-1},\boldsymbol{x}^{t}).
\label{eq:tvaerecurrent}
\end{equation}
$\text{GRU}(*)$ is the calculation of hidden states in a GRU unit. GRU adopts gates and memory cells and alleviates the gradient vanishing problem while being easier to train than LSTM due to fewer gates used. The structure of GRU is formulated as follows:
\begin{equation}
\begin{split}
\boldsymbol{r}^{t}&=\sigma(W_{r}[\boldsymbol{h}^{t-1};\boldsymbol{x}^{t}]),\\
\boldsymbol{\zeta}_{t}&=\sigma(W_{\zeta}[\boldsymbol{h}^{t-1};\boldsymbol{x}^{t}]),\\
\tilde{\boldsymbol{h}}^{t}&=tanh(W_{\tilde{h}}[\boldsymbol{r}^{t}\circ\boldsymbol{h}^{t-1};\boldsymbol{x}^{t}]),\\
\boldsymbol{h}^{t}&=(\boldsymbol{1}-\boldsymbol{\zeta}_{t})\circ\boldsymbol{h}^{t-1}+\boldsymbol{\zeta}_{t}\circ\tilde{\boldsymbol{h}}^{t},
\end{split}
\label{eq:gru}
\end{equation}
where $\circ$ is the element-wise product. Specifically, $\boldsymbol{r}^{t}$ and $\boldsymbol{\zeta}^{t}$ are the reset gate vector and update gate vector, which decides how much past information needs to be forgotten/preserved, respectively. Meanwhile, $\tilde{\boldsymbol{h}}^{t}$ is the candidate activation vector that memorizes the past information, and $\boldsymbol{h}^{t}$ is obtained as the balanced sum of the short ($\boldsymbol{h}^{t-1}$) memory and the long ($\tilde{\boldsymbol{h}}^{t}$) memory.

For the generative process $p(\boldsymbol{x}^{\leq T},\boldsymbol{z}^{\leq T})$, we explicitly simplify $p(\boldsymbol{x}^{t}|\boldsymbol{z}^{\leq t})$ as $p(\boldsymbol{x}^{t}|\boldsymbol{z}^{t})$ to ensure the local pattern of $\boldsymbol{x}^{t}$ is mainly preserved in $\boldsymbol{z}^{t}$; this simplification also can largely reduce the complexity of the reconstruction/decoding process. Based on (\ref{eq:tvaepriortmp}), the generation model can be factorized as follows:
\begin{equation}
p(\boldsymbol{x}^{\leq T},\boldsymbol{z}^{\leq T})=\prod_{t=1}^{T}p(\boldsymbol{x}^{t}|\boldsymbol{z}^{t},\boldsymbol{x}^{<t})p(\boldsymbol{z}^{t}|\boldsymbol{x}^{<t},\boldsymbol{z}^{t-1}),
\label{eq:tvaegeneration}
\end{equation}
$\boldsymbol{p(x}^{t}|\boldsymbol{z}^{t},\boldsymbol{x}^{<t})$ also can be denoted as $p(\boldsymbol{x}^{t}|\boldsymbol{z}^{t},\boldsymbol{h}^{t-1})$, due to the recursive nature of GRU, which requires $\boldsymbol{h}^{t-1}$ being obtained by the recursive calculation with $\boldsymbol{x}^{<t}$.

Similarly, we derive the inference model as:
\begin{equation}
\begin{split}
q(\boldsymbol{z}^{t}|\boldsymbol{x}^{\leq t},\boldsymbol{z}^{t-1})=&\mathcal{N}(\boldsymbol{z}^{t}|\boldsymbol{\mu}_{i}^{t}(\boldsymbol{x}^{\leq t},\boldsymbol{z}^{t-1}),\boldsymbol{\sigma}_{i}^{t}(\boldsymbol{x}^{\leq t},\boldsymbol{z}^{t-1}))
\\
=&\mathcal{N}(\boldsymbol{z}^{t}|\boldsymbol{\mu}_{i}^{t}(\boldsymbol{h}^{t-1},\boldsymbol{x}^{t},\boldsymbol{z}^{t-1}),\boldsymbol{\sigma}_{i}^{t}(\boldsymbol{h}^{t-1},\boldsymbol{x}^{t},\boldsymbol{z}^{t-1})).
\end{split}
\label{eq:tvaeinference}
\end{equation}
The above approximated posterior of $\boldsymbol{z}^{t}$ captures the long-term dynamics carried by $\boldsymbol{x}^{<t}$ ($\boldsymbol{h}^{t-1}$), the neighbouring dependency with $\boldsymbol{z}^{t-1}$, and the corresponding subsequence $\boldsymbol{x}^{t}$.

\subsection{Integration and joint learning}
Based on the encoding of a subsequence and the encoding of the entire time series (represented as subsequences) discussed above, we now integrate them into a HyVAE model, which can jointly learn the local patterns and temporal dynamics for time series forecasting. The jointly learned latent random variables for both time series subsequences and the entire time series are denoted as $\{(\boldsymbol{z}_{L}^{1},...,\boldsymbol{z}_{1}^{1}),...,(\boldsymbol{z}_{L}^{T},...,\boldsymbol{z}_{1}^{T})\}$, with respect to time series $\{\boldsymbol{x}^{1},...,\boldsymbol{x}^{T}\}$, and the encoding process is illustrated in Fig. \ref{fig:rvae}. By combining the prior of subsequence encoding in Eq. (\ref{eq:laveprior}) and the prior of entire time series encoding in Eq. (\ref{eq:tvaepriortmp}), we obtain the prior of HyVAE, which is factorized as follows:
\begin{equation}
\begin{split}
p(\boldsymbol{z}^{t}|\boldsymbol{z}^{t-1},\boldsymbol{h}^{t-1})&=p(\boldsymbol{z}_{L}^{t}|\boldsymbol{z}_{1}^{t-1},\boldsymbol{h}^{t-1})\prod_{i=1}^{L}p(\boldsymbol{z}_{i}^{t}|\boldsymbol{z}_{i+1}^{t}),
\\
p(\boldsymbol{z}_{L}^{t}|\boldsymbol{z}_{1}^{t-1},\boldsymbol{h}^{t-1})&=\mathcal{N}(\boldsymbol{z}_{L}^{t}|\boldsymbol{\mu}^{t}(\boldsymbol{z}_{1}^{t-1},\boldsymbol{h}^{t-1}),\boldsymbol{\sigma}^{t}(\boldsymbol{z}_{1}^{t-1},\boldsymbol{h}^{t-1})),
\\
p(\boldsymbol{z}_{i}^{t}|\boldsymbol{z}_{i+1}^{t})&=\mathcal{N}(\boldsymbol{x}^{t}|\boldsymbol{\mu}_{i}^{t}(\boldsymbol{z}^{t}_{1}),\boldsymbol{\sigma}_{i}^{t}(\boldsymbol{z}^{t}_{1})).
\end{split}
\label{eq:newprior}
\end{equation}
As shown in Fig. \ref{fig:rvae} (a), the prior of HyVAE integrates the long-term temporal dynamics by affecting the first latent random variable of each subsequence (e.g., $\boldsymbol{z}_{L}^{t}$) with the hidden states (e.g., $\boldsymbol{h}^{t-1}$, generated by GRU) of its precedent subsequence. Meanwhile, $\boldsymbol{h}^{t}$ is obtained by the recurrence process with GRU as shown in Fig. \ref{fig:rvae} (b). We then obtain the inference model of HyVAE by integrating Eq. (\ref{eq:laveinference}) and Eq. (\ref{eq:tvaeinference}) as follows (Fig. \ref{fig:rvae} (c)):
\begin{equation}
\begin{split}
q(\boldsymbol{z}^{t}|\boldsymbol{x}^{\leq t},\boldsymbol{z}^{t-1})&=q(\boldsymbol{z}_{L}^{t}|\boldsymbol{x}^{\leq t},\boldsymbol{z}_{1}^{t-1})\prod_{i=1}^{L-1}q(\boldsymbol{z}_{i}^{t}|\boldsymbol{z}_{i+1}^{t},\boldsymbol{x}^{t}),
\\
q(\boldsymbol{z}_{L}^{t}|\boldsymbol{x}^{\leq t},\boldsymbol{z}_{1}^{t-1})&=\mathcal{N}(\boldsymbol{z}_{L}^{t}|\boldsymbol{\mu}^{t}(\boldsymbol{z}_{1}^{t-1},\boldsymbol{x}^{\leq t}),\boldsymbol{\sigma}^{t}(\boldsymbol{z}_{1}^{t-1},\boldsymbol{x}^{\leq t})),
\\
q(\boldsymbol{z}_{i}^{t}|\boldsymbol{z}_{i+1}^{t},\boldsymbol{x}^{t})&=\mathcal{N}(\boldsymbol{z}_{L}^{t}|\boldsymbol{\mu}_{i}^{t}(\boldsymbol{z}_{i+1}^{t},\boldsymbol{x}^{t}),\boldsymbol{\sigma}_{i}^{t}(\boldsymbol{z}_{i+1}^{t},\boldsymbol{x}^{t})).
\end{split}
\label{eq:newinference}
\end{equation}
Similarly, for $\boldsymbol{x}^{t}$, the above encoding process also includes the temporal dynamics carried by $\boldsymbol{h}^{t-1}$ during the encoding of $\boldsymbol{z}_{L}^{t}$, while the rest latent random variables of $\boldsymbol{z}^{t}$ only learn from $\boldsymbol{x}^{t}$. Then, as shown in Fig. \ref{fig:rvae} (d), the generation model of HyVAE is obtained by combining Eq. (\ref{eq:lavegenetion}) and Eq. (\ref{eq:tvaegeneration}) as follows:
\begin{equation}
\begin{split}
p(\boldsymbol{x}^{t}|\boldsymbol{z}^{t},\boldsymbol{x}^{<t})&=p(\boldsymbol{x}^{t}|\boldsymbol{z}_{1}^{t},\boldsymbol{h}^{t-1}),
\\
&=\mathcal{N}(\boldsymbol{x}^{t}|\boldsymbol{\mu}_{i}^{t}(\boldsymbol{z}_{1}^{t},\boldsymbol{h}^{t-1}),\boldsymbol{\sigma}_{i}^{t}(\boldsymbol{z}_{1}^{t},\boldsymbol{h}^{t-1})).
\end{split}
\label{eq:newgeneration}
\end{equation}
Following the derivative process of variational inference, with Eq. (\ref{eq:newprior}) to (\ref{eq:newgeneration}), HyVAE learns the latent representations by maximizing its ELBO defined as follows:
\begin{equation}
\begin{aligned}
\ell_{enc}=&\sum_{t=1}^{T}\Bigg\{\mathbb{E}_{q(\boldsymbol{z}_{l}^{t}|\boldsymbol{x}^{\leq t},\boldsymbol{z}_{1}^{t-1})}\log p(\boldsymbol{x}^{t}|\boldsymbol{h}^{t-1},\boldsymbol{z}_{1}^{t}) \\
  &-\sum_{1}^{l-1}KL\Big(q(\boldsymbol{z}_{i}^{t}|\boldsymbol{z}_{i+1}^{t},\boldsymbol{x}^{t})||p(\boldsymbol{z}_{i}^{t}|\boldsymbol{z}_{i+1}^{t})\Big) \\
  &-KL\Big(q(\boldsymbol{z}_{l}^{t}|\boldsymbol{x}^{\leq t},\boldsymbol{z}_{1}^{t-1})||p(\boldsymbol{z}_{l}^{t}|\boldsymbol{x}^{<t},\boldsymbol{z}_{1}^{<t}\Big)\Bigg\}.
\end{aligned}
\label{eq:HyVAEloss}
\end{equation}
The first term in $\ell_{enc}$ implies the reconstruction loss of HyVAE for each time series subsequence, i.e., between the input $\boldsymbol{x}^{t}$ and the $\hat{\boldsymbol{x}}^{t}$ reconstructed with $(\boldsymbol{z}^{t}_{1},\boldsymbol{h}^{t-1})$ (see Fig. \ref{fig:rvae} (d)). The second and third terms are regularization terms that enforce the encoded latent random variables to jointly capture the local patterns of individual subsequences and learn the temporal dynamics of the entire time series. The expectation of $\ell_{enc}$ is approximated by Monte Carlo estimation \cite{dssmf} and is estimated with the average of the $\ell_{enc}$ of each sample time series.

We use $\boldsymbol{h}^{t}$ and $\boldsymbol{z}^{t}$ for the final time series forecasting, i.e., $\boldsymbol{\hat{y}}=\psi(\boldsymbol{h}^{t},\boldsymbol{z}^{t})$, where $\psi(*)$ is a single-layer fully-connected neural network. The forecasting loss is measured by Eq. (\ref{eq:err}):
\begin{equation}
\ell_{pred}=Err(\boldsymbol{y},\boldsymbol{\hat{y}}).
\label{eq:predictloss}
\end{equation}
Then, the overall loss minimizes the negative ELBO of HyVAE and the forecasting loss as follows:
\begin{equation}
\ell=-\ell_{enc}+\ell_{pred}.
\label{eq:loss}
\end{equation}
In $\ell$, $\ell_{enc}$ aims at learning representations that capture the latent distribution of time series, while $\ell_{pred}$ can be regarded as a regularization term that ensures the latent representations can provide insights for accurate forecasting.
We perform ADMA \cite{adma} for the optimization and use the reparameterization trick \cite{vae} for the model training. For $\ell_{enc}$, we adopt the warm-up scheme \cite{lvae} during the implementation to avoid inactive latent random variables caused by the variational regularization.

\section{Evaluation}
\label{sect:5}
In this section, we first introduce the real-world datasets used to evaluate the proposed method. Then, we explain the accuracy metrics for time series forecasting and briefly describe the counterpart methods. Finally, we analyze the results and compare HyVAE with counterpart methods regarding the effectiveness of time series forecasting. All the experiments are implemented with Python 3.7 and run on a Linux platform with a 2.6G CPU and 132G RAM.

\begin{table}[h]
\centering
\caption{Statistics of the datasets.}
\begin{tabular}{lcccc}
\toprule 
\textbf{Dataset} & \textbf{Train} & \textbf{Valid} & \textbf{Test} &Description\\
\midrule
Parking	      &2856 &357 &358 &Car park occupancy\\  
Stock         &1081 &135 &136 &NASDAQ stock index\\
Electricity	  &1120 &140 &140 &Electricity load values\\ 
Sealevel      &1120 &140 &140 &Sea level pressure\\  
\bottomrule
\end{tabular}
\label{tab:dt}
\end{table}

\subsection{Datasets}
We choose four datasets widely used for time series forecasting. Parking Birmingham dataset \cite{parking} is collected from car parks in Birmingham, which regularly records the total occupancy of all available parking spaces between October 4, 2016, and December 19, 2016. We down-sample the recording frequency to every 5 hours and result in 3571 records. Another NASDAQ stock dataset \cite{nasdaq} consists of stock prices of 104 corporations together with the overall NASDAQ100 index, which is collected from July 26, 2016, to December 22, 2016. We use the NASDAQ100 index for forecasting, and down-sample the records every 30 minutes, which results in 1352 records.
The other two datasets\footnote{https://research.cs.aalto.fi/aml/datasets.shtml.} record the electricity load values of Poland from the 1990s and monthly Darwin sea level pressures from 1882 to 1998, respectively; both datasets contain 1400 records.
We preprocess each dataset with Min-Max normalization by:
\begin{equation}
s'_{i}=\frac{s_{i}-min(\boldsymbol{s})}{max(\boldsymbol{s})-min(\boldsymbol{s})}.
\label{eq:loss}
\end{equation}
Then, each dataset is split into a training set, a validation set and a test set by $\{80\%,10\%,10\%\}$. The number of known time series values used for forecasting is fixed as 50 for all datasets. The statistics of the datasets are shown in Table \ref{tab:dt}.

\begin{table*}[h]
\centering
\caption{Time series forecasting results on the datasets, with the best displayed in bold.}
\begin{tabular}{lccc|ccc|ccc|ccc}
\toprule 
\multirow{2}{*}{\textbf{Methods}} & \multicolumn{3}{c}{\textbf{Parking}} & \multicolumn{3}{c}{\textbf{Stock}} & \multicolumn{3}{c}{\textbf{Electricity}}  & \multicolumn{3}{c}{\textbf{Sealevel}}  \\
\cmidrule{2-4} \cmidrule{5-7} \cmidrule{8-10} \cmidrule{11-13} 
&  \begin{tabular}{@{}c@{}}\textbf{MSE} \\ \tiny{($\times 10^{-2}$)}\end{tabular} &\textbf{MAE} & \textbf{MAPE} & \begin{tabular}{@{}c@{}}\textbf{MSE} \\ \tiny{($\times 10^{-2}$)}\end{tabular} &\textbf{MAE} & \textbf{MAPE} & \begin{tabular}{@{}c@{}}\textbf{MSE} \\ \tiny{($\times 10^{-2}$)}\end{tabular} &\textbf{MAE} & \textbf{MAPE} & \begin{tabular}{@{}c@{}}\textbf{MSE} \\ \tiny{($\times 10^{-2}$)}\end{tabular} &\textbf{MAE} & \textbf{MAPE}\\

\midrule
AR          &1.043 &0.085 &0.474 &1.876 &0.114 &0.146 &1.093 &0.086 &0.336 &1.481 &0.095 &0.187\\
ARIMA       &0.637 &0.066 &0.289 &0.880 &0.080 &0.100 &0.743 &0.056 &0.249 &1.613 &0.107 &0.204\\
SVR         &1.077 &0.082 &0.288 &0.606 &0.075 &0.092 &0.563 &0.057 &0.254 &1.003 &0.079 &0.161\\ \midrule
LSTM        &0.571 &0.057 &0.249 &0.557 &0.068 &0.078 &0.321 &0.036 &0.211 &0.801 &0.068 &0.139\\
Informer &0.425 &0.051 &0.224 &0.728 &0.078 &0.090 &0.305 &0.037 &0.219 &1.083 &0.083 &0.173\\
CNN+LSTM    &0.397 &0.046 &0.200 &0.254 &0.043 &0.049 &0.149 &0.023 &0.210 &0.667 &0.062 &0.127\\ \midrule
Vanilla VAE       &0.713 &0.067 &0.312 &15.142 &0.373 &0.754 &5.445 &0.200 &0.457 &4.386 &0.178 &0.332\\
VRNN       &0.454 &0.056 &0.226 &0.190 &0.039 &0.042 &0.199 &0.029 &0.195 &0.665 &0.066 &0.125\\
LaST       &0.366 &0.043 &0.191 &0.119 &0.026 &0.029 &0.116 &0.018 &0.164 &0.674 &0.064 &0.128\\
HyVAE       &\textbf{0.133} &\textbf{0.028} &\textbf{0.144} &\textbf{0.087} &\textbf{0.021} &\textbf{0.023} &\textbf{0.097} &\textbf{0.015} &\textbf{0.143} &\textbf{0.623} &\textbf{0.060} &\textbf{0.123}\\
\bottomrule
\end{tabular}
\label{tab:acc}
\end{table*}

\begin{table*}[h]
\centering
\caption{Multi-step forecasting results (MSE$\times 10^{-2}$) on the datasets, with the best displayed in bold.}
\begin{tabular}{lccc|ccc|ccc|ccc}
\toprule
\multirow{2}{*}{\textbf{Methods}} & \multicolumn{3}{c}{\textbf{Parking}} & \multicolumn{3}{c}{\textbf{Stock}} & \multicolumn{3}{c}{\textbf{Electricity}}  & \multicolumn{3}{c}{\textbf{Sealevel}}  \\
\cmidrule{2-4} \cmidrule{5-7} \cmidrule{8-10} \cmidrule{11-13} 
&  3 & 4 & 5 & 3 & 4 & 5 & 3 & 4 & 5 & 3 & 4 & 5\\
\midrule
LSTM        &0.712 &0.696 &0.743 &0.669 &0.778 &1.388 &0.340 &0.396 &0.386 &0.938 &1.052 &1.122\\   
Informer &0.662 &0.638 &0.649 &0.933 &1.163 &1.297 &0.307 &0.409 &0.450 &1.105 &1.177 &1.056\\        
CNN+LSTM    &0.635 &0.643 &0.636 &0.524 &0.710 &0.822 &0.163 &0.156 &0.197 &0.890 &0.959 &1.032\\  \midrule
VRNN        &0.614 &0.700 &0.760 &0.557 &0.653 &0.895 &0.232 &0.437 &0.458 &0.870 &1.257 &1.322\\ 
LaST        &\textbf{0.442} &0.533 &0.574 &0.425 &0.463 &0.617 &0.136 &0.138 &0.189 &0.819 &0.892 &1.021\\ 
HyVAE       &0.446 &\textbf{0.466} &\textbf{0.502} &\textbf{0.367} &\textbf{0.365} &\textbf{0.431}  &\textbf{0.123} &\textbf{0.137} &\textbf{0.151} &\textbf{0.787} &\textbf{0.858} &\textbf{0.931}\\
\bottomrule
\end{tabular}
\label{tab:multi}
\end{table*}

\subsection{Performance metric}
We use three different metrics widely used for time series forecasting \cite{cnnlstm,vrnn} in the evaluation, and they are mean square error (MSE), mean absolute error (MAE), and mean absolute percentage error (MAPE).
MSE and MAE respectively measure the variance and average of the residuals of the forecasting results to ground truth and are respectively defined as follows:
\begin{equation}
MSE=\frac{1}{n}\sum_{i=1}^{n}(\boldsymbol{y}_{i}-\hat{\boldsymbol{y}}_{i})^{2},
MAE=\frac{1}{n}\sum_{i=1}^{n}|\boldsymbol{y}_{i}-\hat{\boldsymbol{y}}_{i}|.
\label{eq:mae}
\end{equation}
MAPE measures the proportion of forecasting deviation to the ground truth as follows:
\begin{equation}
MAPE=\frac{1}{n}\sum_{i=1}^{n}|\frac{\boldsymbol{y}_{i}-\hat{\boldsymbol{y}}_{i}}{\max(\epsilon,\boldsymbol{y}_{i})}|,
\label{eq:mape}
\end{equation}
where $\epsilon$ is an arbitrarily small positive value to ensure the dividing is always legal.

\subsection{Counterpart methods}
We select three types of counterpart methods to compare with the proposed method, i.e., the classical statistical models, deterministic DNN-based methods, and VAE-based methods. The classical models include the widely used AR, ARIMA, and SVR. For deterministic DNN-based methods, we choose the LSTM and Informer and implement a stacked CNN and LSTM model (CNN+LSTM) following \cite{cnnlstm} for time series forecasting. For the VAE-based methods, other than the vanilla VAE, we adopt VRNN \cite{vrnn} and LaST \cite{last}. We brief these methods as follows:

\begin{itemize}
\item[--] \textbf{AR} forecasts with the weighted sum of past values. \textbf{ARIMA} incorporates moving average and differencing to AR for non-stationary time series. 
\item[--] \textbf{SVR} \cite{svr} is based on the support vector machine (SVM) and the principle of structural risk minimization.
\item[--] \textbf{LSTM} \cite{lstm} is an RNN model that can learn the long dynamics with its forget gates. 
\item[--] \textbf{Informer} \cite{transformer} uses multi-head attention with position encoding to learn the latent structure of time series for forecasting. 
\item[--] \textbf{CNN+LSTM} \cite{cnnlstm} stacks CNN and LSTM for accurate air quality forecasting. CNN+LSTM includes three TCN layers and two bi-LSTM layers. 
\item[--] \textbf{Vanilla VAE} \cite{vae} is the basic variational autoencoder that learns latent representations as independent Gaussian random variables. 
\item[--] \textbf{VRNN} \cite{vrnn} extends VAE to be capable of learning temporal dynamics by introducing temporal dependency among the latent representations. 
\item[--] \textbf{LaST} \cite{last} adopts disentangled variational autoencoder to capture seasonality and trend, with auxiliary objectives to ensure dissociate representations.
\end{itemize}

\subsection{Experiment setup}

In all the experiments, we use the validation sets to tune optimal parameters and use the test sets for forecasting accuracy measurement. For AR, we search the optimal number of $lag$ (past time series values) from 1 to 10, and use the same strategy to search optimal $p$ (the number of past observations) and $q$ (the size of moving average window) for ARIMA, with the optimal differencing degree searched from 0 to 3. For SVR, we adopt the radial basis function ($RBF$) kernel for running, with its parameters $C$ (regularization parameter) searched from $\{1,10,100,1000\}$ and $\gamma$ (kernel coefficient) searched from $\{0.00005,0.0005,0.005,0.05\}$. 

For LSTM, Informer, CNN+LSTM, vanilla VAE, VRNN, LaST, and HyVAE, we search the optimal batch size from $\{32,64,128\}$ and set the maximum iteration to be 100 epochs. The learning rate is searched from $\{0.001,0.01,0.1\}$.
The dimension of the LSTM/GRU hidden states and latent representations are searched from $\{8,16,32,64,128\}$, and the number of layers is no more than 3. For HyVAE, the $ladder\ size$ and the $subsequence\ length$ are searched from $\{2,4,6,8,10\}$ and $\{10,20,30,40\}$, respectively. We run each method 50 times and report the average accuracy as the final results.

\begin{figure*}[!htbp]
\centering
\subfloat[Parking]{\includegraphics[width=3.46in]{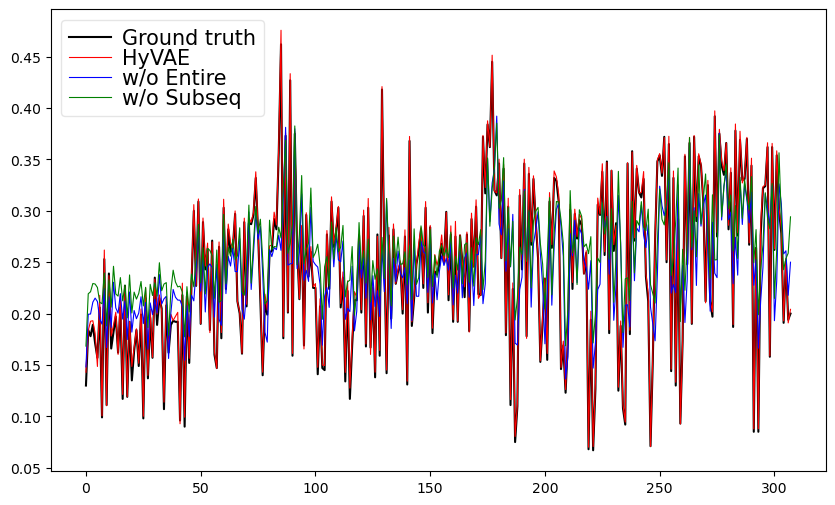}}
\subfloat[Stock]{\includegraphics[width=3.41in]{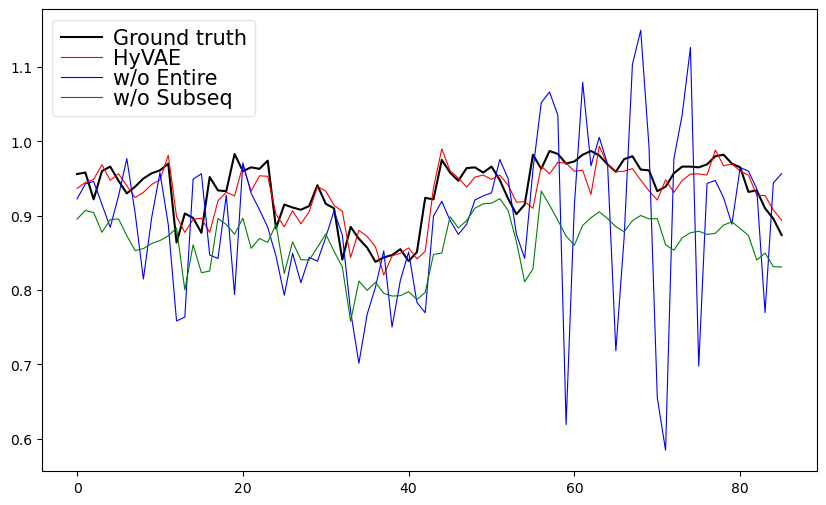}} \\
\subfloat[Electricity]{\includegraphics[width=3.46in]{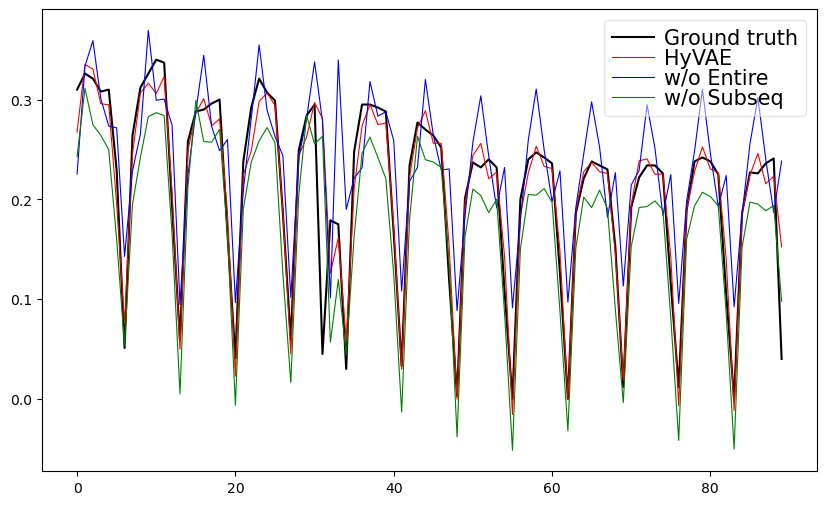}}
\subfloat[Sealevel]{\includegraphics[width=3.41in]{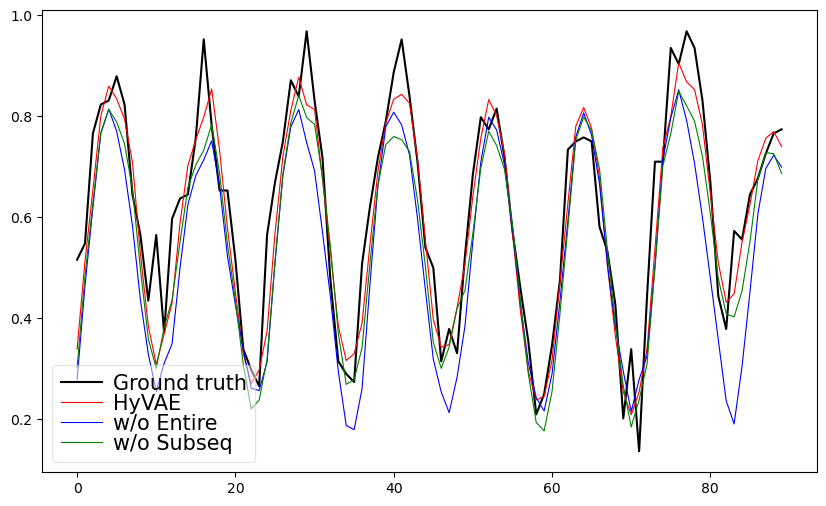}} 
\caption{Forecasting results of HyVAE, w/o Entire and w/o Subseq on the datasets.}
\label{fig:ablation}
\end{figure*}

\subsection{Main results}
In this experiment, we compare the accuracy of HyVAE with counterpart methods, with respect to single-step forecasting and multi-step (3, 4, and 5 steps) forecasting, respectively, on the four datasets. 

As shown in Table \ref{tab:acc}, HyVAE generally achieves the best performance among all methods on the four datasets. Notably, we observe on the Parking dataset, the MSE achieved by HyVAE ($0.133\times 10^{-2}$) is nearly three times smaller than that of the second-best performed LaST ($0.366\times 10^{-2}$). The least improvement over all the counterpart methods is shown in the Sealevel dataset, in which HyVAE reduces the MSE, MAE, and MAPE of VRNN (the second best) by $6.3\%$, $9.1\%$ and $1.6\%$, respectively. When further considering the type of the counterpart methods, first, we see HyVAE achieves significant improvement over the classical AR, ARIMA, and SVR methods, by achieving nearly one magnitude smaller MSE on the Parking and Stock datasets. Second, compared with the deterministic DNN-based LSTM, Informer, and CNN+LSTM, HyVAE also shows significant improvement; especially on the Stock dataset, HyVAE achieves around two times smaller MSE, MAE, and MAPE than the best-performed deep neural network model (CNN+LSTM). Although CNN+LSTM also considers both local patterns and temporal dynamics of time series and outperforms LSTM and Informer on all the datasets, HyVAE constantly being more effective and thus is better at capturing the complex structure of time series for forecasting. Third, we can see that HyVAE achieves more accurate forecasting results than other VAE-based methods that only learn part of the information of time series. That includes the vanilla VAE, which misses the temporal dynamics, VRNN, which only learns the temporal dynamics, and LaST for seasonality/trend patterns of time series. This observation shows the effectiveness of HyVAE to learn both the local patterns and temporal dynamics for time series forecasting. 

For multi-step forecasting, in Table \ref{tab:multi}, we show the MSE of LSTM, Informer, CNN+LSTM, VRNN, LaST, and HyVAE; AR, ARIMA, SVR, and vanilla VAE are excluded due to low performance. The results show that HyVAE achieves more accurate forecasting results than compared counterpart methods. Although generally, the forecasting accuracy decreases with larger forecasting steps, except for the Parking dataset, the forecasting accuracy of HyVAE decreases much slower than the compared counterpart methods since it captures more informative patterns of time series. For example, from 3-step forecasting to 5-step forecasting in the Electricity dataset, the MSE of HyVAE only decreases by 0.028, while LSTM, Informer, CNN+LSTM, VRNN, and LaST decrease by 0.046, 0.143, 0.034, 0.226 and 0.063, respectively. Meanwhile, CNN+LSTM and HyVAE constantly produce more accurate forecasting results than other deterministic DNN-based methods and VAE-based methods, respectively, and that again supports the effectiveness of learning both local patterns and temporal dynamics for time series forecasting.

\subsection{Ablation analysis}
We conduct an ablation analysis to further understand the effectiveness of learning both the local patterns and the temporal dynamics in HyVAE. To do that, we implement two variants of HyVAE by removing the learning of one type of information, respectively; that is, w/o Subseq that excludes the learning of local patterns from subsequences, and w/o Entire that does not learn the temporal dynamics of the entire time series. The parameters of w/o Subseq and w/o Entire are tuned with the validation set the same as HyVAE, and we show the results of time series forecasting measured by MSE in Table \ref{tab:ablation}.

\begin{table}[h]
\centering
\caption{Ablation analysis of HyVAE (MSE$\times 10^{-2}$), with the best displayed in bold.}
\begin{tabular}{lcccc}
\toprule
Methods & Parking & Stock & Electricity  & Sealevel  \\
\midrule
w/o Entire  &0.410 &0.980 &0.503 &1.742\\ 
w/o Subseq  &0.389 &0.513 &0.160 &1.088\\ 
HyVAE       &\textbf{0.133} &\textbf{0.087} &\textbf{0.116} &\textbf{0.623}\\
\bottomrule
\end{tabular}
\label{tab:ablation}
\end{table}

In Table \ref{tab:ablation}, HyVAE that learns both information achieves higher forecasting accuracy than the two variants. Specifically, the largest improvement of HyVAE towards the variants is shown in the Stock dataset (0.087), i.e., around six times smaller than that of w/o Subseq (0.513, second best). The smallest improvement appears in the Sealevel dataset, but the MSE of HyVAE is still around two times smaller than that of the second-best performed w/o Subseq (0.623 to 1.088). Meanwhile, it is interesting to observe that w/o Entire, which misses the temporal dynamics, constantly performs worse that w/o Subseq, which misses local patterns, on the four datasets. 

\begin{figure*}[!htbp]
\centering
\subfloat[Parking]{\includegraphics[width=1.69in]{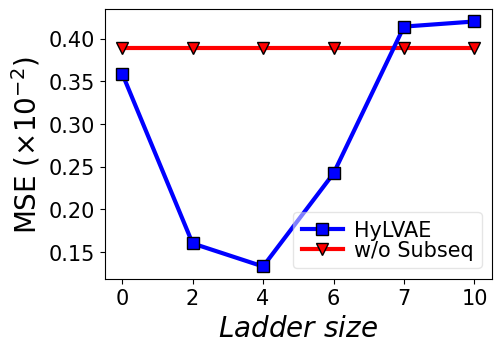}}
\subfloat[Stock]{\includegraphics[width=1.65in]{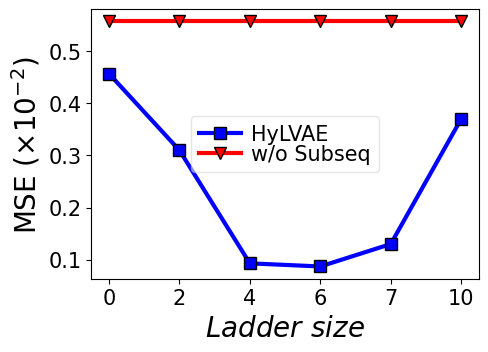}} 
\subfloat[Electricity]{\includegraphics[width=1.69in]{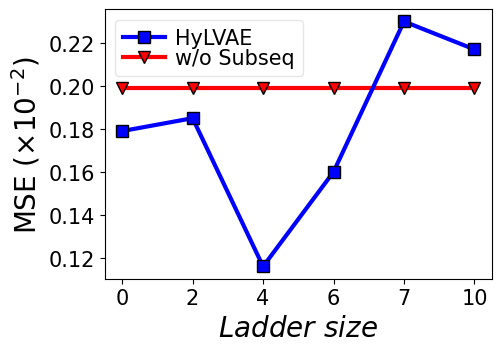}}
\subfloat[Sealevel]{\includegraphics[width=1.65in]{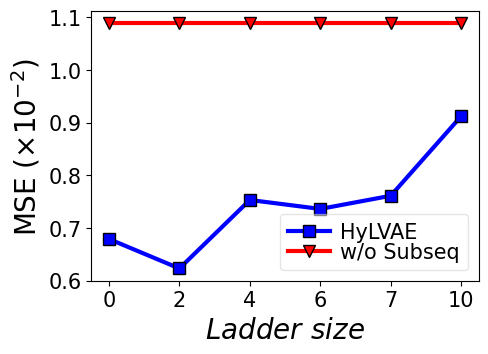}} \\
\subfloat[Parking]{\includegraphics[width=1.69in]{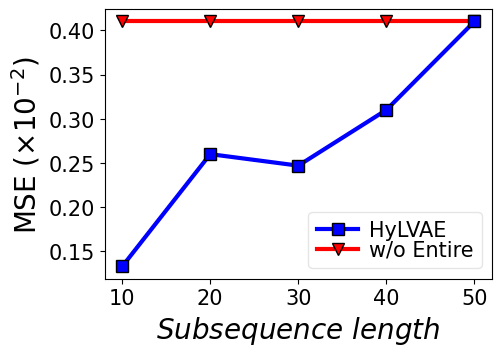}}
\subfloat[Stock]{\includegraphics[width=1.65in]{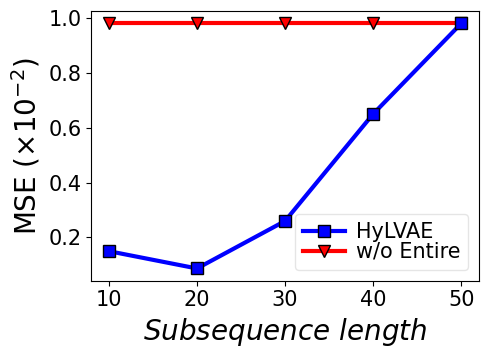}} 
\subfloat[Electricity]{\includegraphics[width=1.69in]{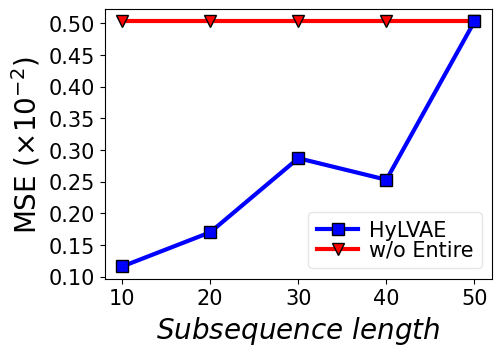}}
\subfloat[Sealevel]{\includegraphics[width=1.65in]{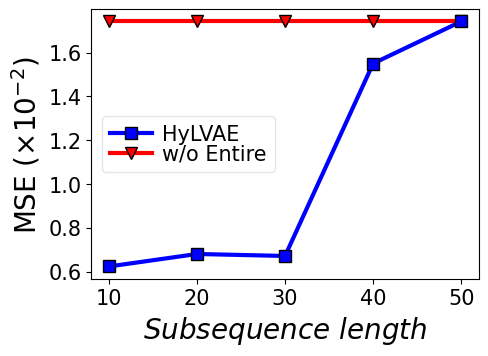}} \\
\caption{Parameter analysis of HyVAE with respect to $ladder\ size$ and $subsequence\ length$.}
\label{fig:param1}
\end{figure*}

\begin{figure*}[!htbp]
\centering
\subfloat[Parking]{\includegraphics[width=1.6in]{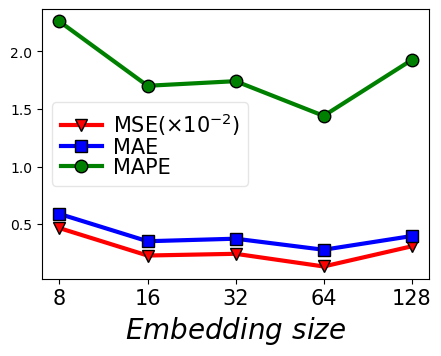}}\hspace{0.1in}
\subfloat[Stock]{\includegraphics[width=1.6in]{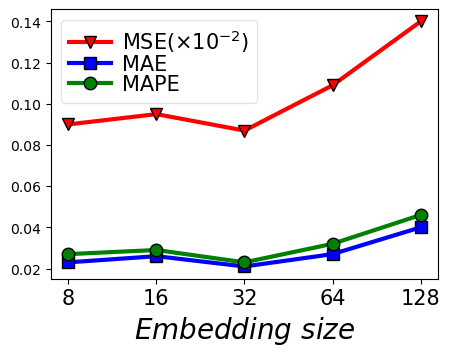}} 
\subfloat[Electricity]{\includegraphics[width=1.6in]{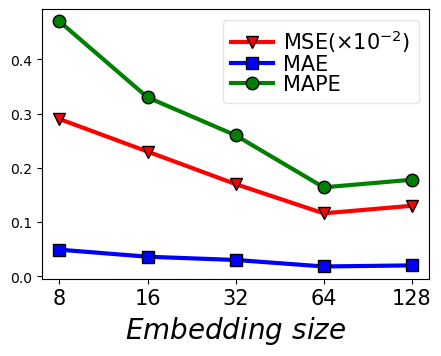}}\hspace{0.1in}
\subfloat[Sealevel]{\includegraphics[width=1.6in]{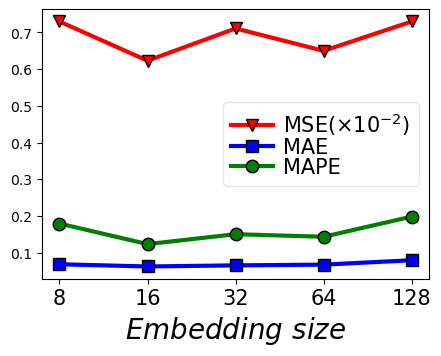}} 
\caption{Parameter analysis of HyVAE with respect to $embedding\ size$.}
\label{fig:param2}
\end{figure*}

We further show the forecasting results of HyVAE and the variants against the ground truth in Fig. \ref{fig:ablation}. For w/o Entire, since no temporal dynamics is learned, it cannot properly capture the global trend of time series; especially for the Stock dataset (Fig. \ref{fig:ablation} (b)), it misunderstands the steady curves between the 55 and 75 timestamps as sharp spikes. As for the results of the Electricity dataset and the Sealevel dataset shown in Fig. \ref{fig:ablation} (c-d), w/o Entire only emphasizes recurring local patterns but misses their differences at different timestamps. Meanwhile, w/o Subseq can better express the temporal dynamics than w/o Entire, as clearly shown in Fig. \ref{fig:ablation} (b); however, it fails to properly capture local details. By combining the strengths of w/o Entire and w/o Subseq, HyVAE achieves the best forecasting results that are quite close to the ground truth.

\subsection{Parameter analysis}
In this experiment, we analyze the impact of three parameters of HyVAE, i.e., the $ladder\ size$, the $subsequence$ $length$, and the $embedding\ size$, on its performance. Specifically, the $ladder\ size$ determines the causal information during subsequence encoding, and we vary the $ladder\ size$ from 0 to 10, where 0 means HyVAE learns no causal information of subsequences (see Fig. \ref{fig:lvae}). The $subsequence\ length$ balances the local patterns and the temporal dynamics, i.e., HyVAE is degraded to w/o Subseq or w/o Entire if $subsequence\ length$ is 0 (no subsequence) or the maximum (50, the subsequence becomes the entire time series), respectively.    
The results measured by MSE are shown in Fig. \ref{fig:param1}. For the $ladder\ size$ shown in Fig. \ref{fig:param1} (a-d), the forecasting accuracy significantly decreases when $ladder\ size$ is too small or too large. The optimal ladder sizes are relatively small (2 for the Sealevel dataset, 4 for the Parking and Stock datasets, and 6 for the Electricity dataset). Meanwhile, we see that when the $ladder\ size$ equals 0, HyVAE still outperforms w/o Subseq on all the datasets. The results of $subsequence\ length$ are shown in Fig. \ref{fig:param1} (c-d), in which we see that HyVAE prefers short subsequences to obtain optimal forecasting results, i.e., $subsequence\ length$ is 10 (Parking, Electricity, and Sealevel datasets) or 20 (Stock dataset). The reason is that if the subsequences are too long, temporal dynamics can hardly be preserved. Not surprisingly, HyVAE that learns temporal dynamics with different $subsequence\ length$ achieves better forecasting accuracy than w/o Entire, which does not learn temporal dynamics at all.

We then run HyVAE with varying $embedding\ size$ $\{8,16,32,64,128\}$, which determines the dimension of latent representation and the dimension of hidden states in neural networks, and the results are shown in Fig. \ref{fig:param2}. On all the datasets, accuracy measured by MSE, MAE, and MAPE has similar trends. First, the forecasting accuracy is low with small $embedding\ size$, mainly because the small size of latent random variables cannot properly capture the complex non-linear processes of time series. When the $embedding\ size$ becomes too large (128), the accuracy decreases due to over-fitting. The results show that HyVAE can obtain optimal forecasting results with relatively small $embedding\ size$.

\section{Conclusion}
\label{sect:6}
This paper proposes a novel hybrid variational autoencoder (HyVAE) model for time series forecasting. HyVAE integrates the learning of local patterns and temporal dynamics into a variational autoencoder. Through comprehensive evaluation on four real-world datasets, we show that HyVAE achieves better time series forecasting accuracy than various counterpart methods, including a deterministic DNN-based method (CNN+LSTM) that also learns both information of time series.
Moreover, the ablation analyses demonstrate that HyVAE outperforms its two variants which only learn local patterns or temporal dynamics from time series.

\bibliographystyle{elsarticle-num}

\bibliography{ref}

\end{document}